\documentclass{llncs}

\usepackage{epsfig}
\usepackage{amsmath}
\usepackage{amssymb}
\usepackage{epic, eepic}
\usepackage{latexsym}
\usepackage{subfigure}
\usepackage[ruled,vlined]{algorithm2e}
\usepackage{color}

\newcommand{\set}[1]{\{ #1 \}}

\newcommand{\pr}{{\mathrm{Pr}}}
\newcommand{\params}[1]{\langle #1 \rangle}
\newcommand{\mb}[1]{\mathbf{#1}}
\newcommand{\varPose}{\mb{x}}

\newcommand{\variSAM}{\tau}

\newcommand{\norma}[1]{\left\| #1 \right\|}
\newcommand{\etal}{\textit{et~al.}}
\newcommand{\lcl}[1]{\multicolumn{1}{|c|}{#1}}
\DeclareMathOperator*{\argmax}{arg\,max}
\DeclareMathOperator*{\argmin}{arg\,min}

\begin{document}

\title{Line Maps in Cluttered Environments}

\author{Leonardo Romero and Carlos Lara}

\institute{Division de Estudios de Posgrado\\ Facultad de Ingenieria Electrica\\ %
        Universidad Michoacana.\\ Ciudad Universitaria. 58060, Morelia, Mexico}

\maketitle

\begin{abstract}
This paper uses the smoothing and mapping framework to solve the SLAM problem in indoor environments; focusing on how some key issues such as feature extraction and data association can be handled by applying probabilistic techniques.
For feature extraction, an odds ratio approach to find multiple lines from laser scans is proposed, this criterion allows to decide which model must be merged and to output the best number of models. In addition, to solve the data association problem a  method based on the segments of each line is proposed. Experimental results show that high quality indoor maps can be obtained from noisy data.
\end{abstract}

\section{Introduction}
The Simultaneous Localization And Mapping (SLAM) is a classical problem in robotics. To solve it, a robot must use the measurements provided by its sensors to estimate a map of the environment and, at the same time, to localize itself within the map. While localization with a given map or mapping with known positions is relatively easy, the combined problem is hard to solve.

Several approaches for SLAM have been proposed, the most successful are based on probabilistic techniques\cite{thrunProbabilistic}. The principal difference between them are the map representation and the uncertainty description; feature--based SLAM approach uses a collection of primitives to represent the map. Features are recognizable structures of elements in the environment. The feature extraction process reduces the complexity by capturing only the essential infomation of the raw data; this data reduction allows to represent and manage the sensor information efficiently but the resulting maps are usually less expressive and precise than when using the raw data. 

Lines and segments are commonly used to represent indoor environments, and the preferred sensor is the laser scan. Many algorithms are available in the literature for extracting multiple lines from range scans; unfortunately, some of these approaches are \textit{ad hoc} methods that use a distance threshold. One of the contributions of this paper is a new method to find multiple lines from laser scans; the formulation presented focuses on estimating a likelihood ratio on the number of lines that are present in a laser scan. The proposed approach follows the Ockham's razor principle; this principle states that the simplest explanation for some phenomenon is more likely to be accurate than more complicated explanations. Other approaches that follows this principle are the Minimum Description Length (MDL) \cite{GrunMDL07,Ying2010} and the Akaike Information Criterion (AIC) \cite{Akaike:1974}.

Another trouble that faces the feature--based SLAM approach is known as the association problem; this problem consists of associating the newest features with those already stored in the map. This is also a crucial problem, and a good option to solve it is the Joint Compatibility Test (JCT)\cite{neira2001}. In this paper, a validation gate is used within the JCT;
the validation gate uses the segments associated with each line to improve the association.

To solve the SLAM problem, the Smoothing and Mapping (SAM) framework\cite{KaessRD08} is implemented; SAM is a smoothing approach rather than the commonly used filtering one. The essential computational advantage arises due to the smoothing information matrix that remains sparse without the need for any approximations; SAM is based on QR and Cholesky matrix factorizations that greatly speed up the optimization procedure leading to a very efficient algorithm.

The rest of the paper is organized as follow: Section \ref{sec:LineExtractor} introduces the new approach to determine the best number of lines and their parameters; Section \ref{sec:slam} overviews the Smoothing and Mapping approach; Section \ref{sec:Association} focuses on the association problem;  Experimental results are presented in Section \ref{sec:results}; finally, Section \ref{sec:conclusions} ends this paper.


\section{Multiple Line Extraction} \label{sec:LineExtractor}

The principal problems to obtain multiple features are: 1) find the best number of features, 2) determine which points belong with each feature, and 3) estimate the feature parameters given its points.  There are many techniques for multiple geometric feature extraction: Bottom-up \cite{duda1973pca,Borges2000,Siegwart}, Probabilistic Techniques \cite{Thrun03IEEETRANS,Han2004}, voting schemes such as Ransac \cite{BollesF81,schnabel-2007-efficient}, Hough Transform \cite{HTSurvey}, etc.

Bottom--up approach has been used in many pattern recognition tasks. At first, local features are extracted from the data. Normally, a distance between each pair of current clusters is computed and the closest pair is merged iteratively until a stopping criterion is met; the simplest criterion involves a threshold: when the distance of the closest features is greater than the predefined threshold the process is stopped \cite{Siegwart}. Here, the result depends strongly on the threshold and the distance used  (usually a Euclidean or  Mahalanobis distance).

Robust regression techniques can manage data with a large proportion of data points that do not belong to the main model. The most widely used robust algorithm is the Random Sampling and Consensus (Ransac) \cite{BollesF81}. Due to its greedy nature, the sequential approach does not consider the relationship among different models; consequently, the result is usually imprecise for non-trivial cases.

To estimate the number of features and their parameters, Thrun \etal \cite{Thrun03IEEETRANS} use real-time variant of the expectation maximization (EM) algorithm; their method penalizes models with many features by using an exponential prior probability; the search for the best number of features is interleaved with the running EM algorithm. The search involves a step for creating new features, and another one for removing features, both executed in regular intervals.

Han \etal \cite{Han2004} formulate the problem of finding features from range images in the Bayesian framework, where prior probabilities penalize the number of features. The algorithm simulates Markov chains with both reversible jumps and stochastic diffusions to traverse the solution space; reversible jumps are used to move among subspaces of different dimensions, such as switching feature models and changing the number of features. 

This paper formulates the problem of finding the number of features from a Bayesian viewpoint; specifically, we study the problem of finding a set of lines from a laser scan. Indoor environments are usually rich in planar surfaces, lines are the natural way to represent them. Nguyen \etal \cite{nguyen_2007_a_comparison_of} compare some algorithms to extract lines from laser scans.

\subsection{Problem Statement and Notation} \label{sec:model}

The multiple line extraction problem is stated as: \textit{Given a set of points from a laser scan, find the best number of line segments and their parameters}. The challenge for any line extraction algorithm consists of finding a realistic representation. 

Let $Z = \set{z_1, \ldots, z_N}$ be a set of measurements obtained from a two--dimensional laser scan; the line extraction problem consists of finding the set of lines 

\begin{equation}
\Theta = \set{\theta_1, \ldots, \theta_M};
\end{equation}%
that better represents $Z$; where, $\theta_j$ represents the parameters of a line.

\subsection{Initial Segmentation}

The proposed algorithm to find lines from laser scan is similar to traditional agglomerative clustering. Initial clusters can be obtained by any conventional bottom--up technique (i.e. sliding window \cite{Siegwart} or Iterative End Point Fit (IEPF) \cite{duda1973pca}). The segmentation step finds a set of $M$ linear clusters ${\mathcal{Z}} = \set{Z_1, \ldots, Z_M}$ from $Z$; each linear cluster $Z_i$ is a group of adjacent points that follow a linear model. These clusters are pairwise disjoint, that is $Z_i \cap Z_j = \set{} \mid \forall i \neq j$. To find the resulting line map, similar clusters must be merged; merging phase is usually based on Euclidean or Mahalanobis distance, the following section formulates the merging phase from a probabilistic viewpoint.

\subsection{Odds ratio test (ORT)} \label{sec:ort}
Let $Z_a \subseteq Z$ be a point cluster that follows a linear model with parameters $\theta_a$; it is interesting to find the probability that $Z_a$ is generated by $\theta_a$. Assuming independence among measurements%
\begin{eqnarray} 
\pr (Z_a \mid  \theta_a) &\propto& \prod_{z_i \in Z_a} \exp \left[-\frac{d^2_{\perp}(z_i,  \theta_a)}{2 \sigma^2} \right] = 
\exp \left[-\sum_{z_i \in Z_a} \frac{d^2_{\perp}(z_i,  \theta_a)}{2 \sigma^2} \right] \nonumber \\
& \propto &  \exp \left(- \frac{\chi^2_a}{2} \right),  
\label{eq:probClusterA} 
\end{eqnarray} 
\noindent where  
 
\begin{equation} 
\chi^2_a = \sum_{z_i \in Z_a} \frac{d^2_{\perp}(z_i,  \theta_a)}{\sigma^2}. 
\end{equation}


Using Equation (\ref{eq:probClusterA}) and considering independent gaussian measurements, the likelihood for the $M$ clusters is

\begin{align} 
\pr (Z \mid M, I) & = \frac{1}{ \left( 2\pi r_{\mathrm{max}} \right) ^M } \int \ldots \int  \exp \left( -\frac{1}{2} \sum_{j=1}^M \chi_j^2 \right) d^M r_j d^M \alpha_j \nonumber \\ 
& =  \frac{1}{ \left( 2 \pi r_{\mathrm{max}} \right) ^M } \prod_{j=1}^M \int  \int  e^{-\frac{\chi_j^2}{2}} d r_j d \alpha_j, 
\label{eq:startA}
\end{align} 
 
The $(M-1)$--lines model is obtained by merging two clusters $Z_{a}, Z_{b} \in \mathcal{Z}$ into a new cluster $Z_c = Z_a \cup Z_{b}$. The likelihood in that case is 
 
\begin{align} 
\pr  (Z \mid  & M-1, I) = \frac{1}{ \left( 2 \pi r_{\mathrm{max}} \right) ^{M-1} } \times 
\frac{\int  e^{-\frac{\chi_c^2}{2}} d r_c d \alpha_c \times \prod_{j=1}^M \int \int e^{\frac{-\chi_j^2}{2}} d r_j d \alpha_j}{\int \int  e^{-\frac{\chi_a^2}{2}} d r_a d \alpha_a  \int \int e^{-\frac{\chi_b^2}{2}} d r_b d \alpha_b} 
\end{align} 

To choose between the $M$--lines model and the \mbox{$(M-1)$--lines} model we compare their likelihood by using the ratio: 
\begin{equation} 
\frac{\pr (Z \mid M-1, I)}{\pr (Z \mid M, I)} =  
\frac{2 \pi r_{\mathrm{max}} \int \int  e^ {-\frac{\chi_c^2}{2}} d r_c d \alpha_c} 
{\int  \int e^{ -\frac{\chi_a^2}{2}} d r_a d \alpha_a \int \int  e^{-\frac{\chi_b^2}{2}} d r_b d \alpha_b}. \label{eq:CompEvid} 
\end{equation} 
The integrals of Equation (\ref{eq:CompEvid}) can be solved by using a Taylor series expansion about the parameters of the least square line for each cluster. Let be $\theta_{j*} = \params{r_{j*}, \alpha_{j*}}$ the parameters of the least square line for the $j$--th cluster and $\chi^2_{j*}$ its corresponding error. Taylor expansion about $\chi^2_{j*}$ gives%
\begin{equation} 
\chi^2_j \approx \chi^2_{j*} + \frac{1}{2}(\theta - \theta_{j*})^T  \nabla \nabla \chi^2_j (\theta - \theta_{j*}) + \ldots, \label{eq:taylor} 
\end{equation}%
\noindent where $\nabla \nabla \chi^2 $ is the Hessian matrix, evaluated at $\theta_{j*}$. Finally, the odds ratio is found by solving Equation (\ref{eq:CompEvid}) with Equation (\ref{eq:taylor}):

\begin{align} 
R = \frac{\pr (Z \mid M-1, I)}{\pr (Z \mid M, I)}  = \noindent
\frac{r_{ \mathrm{max}}}{2}  & \sqrt{ \frac{\det(\nabla \nabla \chi^2_a)  \det(\nabla \nabla \chi^2_b)}  
  { \det (\nabla \nabla \chi^2_{c})}} \quad \quad \nonumber \\ 
& \times \exp \left[ \frac{1}{2} (\chi^2_{a*}+\chi^2_{b*}-\chi^2_{c*} )\right].
\label{eq:conclusion} 
\end{align}  
 
When the value of the odds ratio (Eq.  \ref{eq:conclusion}) is equal to one, both models are equally like; lower values mean that the $(M-1)$--lines model obtained by merging $Z_a$ and $Z_b$ is less likely to occur than the $M$--model; that is,  clusters $Z_a$ and $Z_b$ should not be merged. On the other hand, values greater than one prefer the $(M-1)$--model. Then, the odds ratio can be used to decide greedily which two clusters to merge. Equation \ref{eq:conclusion} follows the Ockham's razor principle: factors that multiply the exponential term penalizes models with more features.   

\subsection{Proposed Algorithm} \label{sec:proposed} 
 
\begin{algorithm}[!b] 
\KwIn{A laser scan $Z = \set{z_1, \ldots, z_N}$} 
\KwOut{A set of lines $\Theta = \set{\theta_1, \ldots, \theta_M}$ and their corresponding  clusters $\mathcal{Z} = \set{Z_1, \ldots, Z_M}$} 
Find a set of local clusters $\mathcal{Z} = \set{Z_1, \ldots, Z_M}$ where $\bigcup_{i=1}^M Z_i \subseteq Z$, and $Z_i \cap Z_j = \set{}$ for $i \neq j$\; \nllabel{step:initialization} 
 
\Repeat{$r \leq 1 $}{ 
        Find the pair $Z_a, Z_b \in \mathcal{Z}$ with the highest probability of the merged hypothesis \; \nllabel{step:pair} 
        $r \gets \frac{\pr (Z \mid M-1, I)}{\pr (Z \mid M, I)}$ \nllabel{step:r} 
 
         \If{$r>1$}{ \nllabel{step:mergeA} 
                $Z_a \gets Z_a \cup Z_b$ \; 
                $\mathcal{Z} \gets \mathcal{Z} \setminus \set{Z_b}$ \; 
                $M \gets M - 1$ 
         }\nllabel{step:mergeB} 
} 
$\Theta = \set{\theta_j \mid j = 1, \ldots, M}$ where $\theta_j$ is the best line for the cluster $Z_j$ in the least squares sense \; 
\Return{$\Theta, \mathcal{Z}$ } 
\caption{Proposed Algorithm} 
\label{alg:mine} 
\end{algorithm}

As is shown in previous section, pairs of clusters can merged iteratively based on the ratio given by Equation \ref{eq:conclusion}. A tree provides a picture for agglomerative clustering techniques, in this sense the value obtained from Equation \ref{eq:conclusion} also helps to decide the cutting height of the tree. That is, when the value of $R$ is less than or equal to one, the best value for the number of models $M$ has been found.  This approach is valid because the densities involved are MLR (i.e. have a monotone likelihood ratio), and since the Gaussian model is MLR, the odds ratio test is applicable. 

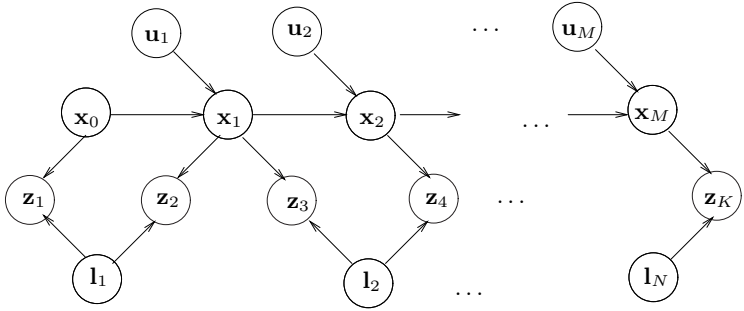
\begin{figure}[!htb]
\centering
\setlength{\unitlength}{0.00049in}
\begingroup\makeatletter\ifx\SetFigFontNFSS\undefined%
\gdef\SetFigFontNFSS#1#2#3#4#5{%
  \reset@font\fontsize{#1}{#2pt}%
  \fontfamily{#3}\fontseries{#4}\fontshape{#5}%
  \selectfont}%
\fi\endgroup%
{\renewcommand{\dashlinestretch}{30}
\begin{picture}(7906,3305)(0,-10)
\put(3078,1155){\ellipse{524}{524}}
\put(270,1162){\ellipse{524}{524}}
\put(1730,1159){\ellipse{524}{524}}
\put(4601,1176){\ellipse{524}{524}}
\put(7636,1205){\ellipse{524}{524}}
\put(990,313){\whiten\ellipse{524}{524}}
\put(990,313){\ellipse{524}{524}}
\put(3902,269){\whiten\ellipse{524}{524}}
\put(3902,269){\ellipse{524}{524}}
\put(6955,331){\whiten\ellipse{524}{524}}
\put(6955,331){\ellipse{524}{524}}
\path(865,547)(416,996)
\path(522.066,932.360)(416.000,996.000)(479.640,889.934)
\path(1158,479)(1607,928)
\path(1543.360,821.934)(1607.000,928.000)(1500.934,864.360)
\path(3722,480)(3273,929)
\path(3379.066,865.360)(3273.000,929.000)(3336.640,822.934)
\path(7143,524)(7592,973)
\path(7528.360,866.934)(7592.000,973.000)(7485.934,909.360)
\path(4083,479)(4532,928)
\path(4468.360,821.934)(4532.000,928.000)(4425.934,864.360)
\put(3137,2969){\ellipse{524}{524}}
\put(1626,2950){\ellipse{524}{524}}
\put(873,2100){\whiten\ellipse{524}{524}}
\put(873,2100){\ellipse{524}{524}}
\put(2396,2076){\whiten\ellipse{524}{524}}
\put(2396,2076){\ellipse{524}{524}}
\put(3926,2068){\whiten\ellipse{524}{524}}
\put(3926,2068){\ellipse{524}{524}}
\put(6139,3021){\ellipse{524}{524}}
\put(6945,2127){\whiten\ellipse{524}{524}}
\put(6945,2127){\ellipse{524}{524}}
\path(1136,2076)(2126,2076)
\path(2006.000,2046.000)(2126.000,2076.000)(2006.000,2106.000)
\path(2666,2076)(3656,2076)
\path(3536.000,2046.000)(3656.000,2076.000)(3536.000,2106.000)
\path(868,1853)(419,1404)
\path(482.640,1510.066)(419.000,1404.000)(525.066,1467.640)
\path(2308,1853)(1859,1404)
\path(1922.640,1510.066)(1859.000,1404.000)(1965.066,1467.640)
\path(3318,2729)(3767,2280)
\path(3660.934,2343.640)(3767.000,2280.000)(3703.360,2386.066)
\path(4105,1852)(4554,1403)
\path(4447.934,1466.640)(4554.000,1403.000)(4490.360,1509.066)
\path(2552,1830)(3001,1381)
\path(2894.934,1444.640)(3001.000,1381.000)(2937.360,1487.066)
\path(1811,2751)(2260,2302)
\path(2153.934,2365.640)(2260.000,2302.000)(2196.360,2408.066)
\path(4241,2076)(4871,2076)
\path(4751.000,2046.000)(4871.000,2076.000)(4751.000,2106.000)
\path(7120,1897)(7569,1448)
\path(7462.934,1511.640)(7569.000,1448.000)(7505.360,1554.066)
\path(6333,2774)(6782,2325)
\path(6675.934,2388.640)(6782.000,2325.000)(6718.360,2431.066)
\path(6041,2076)(6671,2076)
\path(6551.000,2046.000)(6671.000,2076.000)(6551.000,2106.000)
\put(1496,2841){\makebox(0,0)[lb]{$\mb{u}_1$}}
\put(3026,2906){\makebox(0,0)[lb]{$\mb{u}_2$}}
\put(5006,2976){\makebox(0,0)[lb]{$\ldots$}}
\put(5276,1131){\makebox(0,0)[lb]{$\ldots$}}
\put(4826,141){\makebox(0,0)[lb]{$\ldots$}}
\put(746,1946){\makebox(0,0)[lb]{$\mb{x}_0$}}
\put(2276,1946){\makebox(0,0)[lb]{$\mb{x}_1$}}
\put(3806,1946){\makebox(0,0)[lb]{$\mb{x}_2$}}
\put(191,1086){\makebox(0,0)[lb]{$\mb{z}_1$}}
\put(1631,1086){\makebox(0,0)[lb]{$\mb{z}_2$}}
\put(3026,1041){\makebox(0,0)[lb]{$\mb{z}_3$}}
\put(4511,1086){\makebox(0,0)[lb]{$\mb{z}_4$}}
\put(7500,1086){\makebox(0,0)[lb]{$\mb{z}_K$}}
\put(911,231){\makebox(0,0)[lb]{$\mb{l}_1$}}
\put(3836,186){\makebox(0,0)[lb]{$\mb{l}_2$}}
\put(5546,1941){\makebox(0,0)[lb]{$\ldots$}}
\put(5966,2901){\makebox(0,0)[lb]{$\mb{u}_M$}}
\put(6851,231){\makebox(0,0)[lb]{$\mb{l}_N$}}
\put(6756,2006){\makebox(0,0)[lb]{$\mb{x}_M$}}
\end{picture}
}
\caption{Bayesian belief network for a SLAM problem. The objective of SLAM is to localize the robot while simultaneously building a map of the environment. The Full SLAM problem requires that the entire robot trajectory is also determined.}
\label{fig:redB}
\end{figure}

\section{SLAM Framework} \label{sec:slam}
The Full SLAM problem can be formulated by a \textit{belief net} representation as shown in Figure \ref{fig:redB};
here, the iSAM approach \cite{KaessRD08} is used to solve it. 
The trajectory of the robot is denoted by  $X = \left[ \varPose_0, \ldots \varPose_M \right]$, where $\varPose_i$ is the $i$--th pose of the robot. The map of the environment is denoted by the set of landmarks $L = \set{\mb{l}_j \mid j = 1, \ldots, N}$, and the measurement set by $Z = \set{\mb{z}_k \mid k = 1, \ldots K}$. For line maps, $\mb{l}_j$ is a line --we use the algorithm introduced in the previous section to obtain line measurements from laser scans. The joint probability model corresponding to this network is
\begin{equation}
 P(X, L, Z) = \Pr(\varPose_0) \prod_{i=1}^M \Pr(\varPose_i \mid \varPose_{i-1}, u_i) \prod _{k=1}^K \Pr(\mb{z}_k \mid \mb\varPose_{i_k}, \mb{l}_{j_k}). \label{eq:SAMJoint}
\end{equation}

Let us denote  $\variSAM = (X, L)$ the variables we are looking for; the maximum a posteriori (MAP) estimate is obtained by
\begin{align}
\variSAM^{*} &= \argmax_{\variSAM} \Pr (X,L\mid Z) \nonumber \\
&= \argmin_{\variSAM} \left(- \log \Pr (X, L, Z) \right). \label{eq:SAMGralProb}
\end{align}

Assuming gaussian process and measurement models, defined by
%
%
\begin{align}
\Pr(\varPose_i \mid \varPose_{i-1}, \mb{u}_i) \propto & \exp - \frac{1}{2}  \norma{\mb{f}_i (\varPose_{i-1}, \mb{u}_i ) - \varPose_i}_{\Lambda_i}^2 \label{eq:modPose2I}\\
\Pr(\mb{z}_k \mid \varPose_{i_k}, \mb{l}_{j_k} ) \propto & \exp  - \frac{1}{2} \norma{\mb{h}_k (\varPose_{i_k} , \mb{l}_{j_k} ) - \mb{z}_k}_{\Gamma_k}^2 \label{eq:modMed2I}; 
\end{align}%
where $\norma{e}^2_{\Sigma} \equiv e ^T \Sigma ^{-1} e $ is the Mahalanobis distance.

A standard non-linear least squared formulation is obtained by combining equations \ref{eq:modPose2I}, \ref{eq:modMed2I} and  \ref{eq:SAMJoint}, and considering as uniform the initial distribution $\Pr(\mb\varPose_0)$,
\begin{align}
\variSAM^{*} &= \argmin_{\variSAM} \left\{ \sum_{i=1}^M \norma{\mb{f}_i (\varPose_{i-1}, \mb{u}_i ) - \varPose_i}^2_{\Lambda_i} + \sum_{k=1}^K \norma{\mb{h}_k (\varPose_{i_k}, \mb{l}_{j_k} ) - \mb{z}_k}^2_{\Gamma_k} \right\}. \label{eq:SAMasLS}
\end{align}%

Kaess \etal \cite{KaessRD08} propose to solve Equation \ref{eq:SAMasLS} incrementally; their method, known as iSAM,  provides an efficient and exact solution by updating a QR factorization of the naturally sparse smoothing information matrix, therefore recalculating only the matrix entries that actually change. iSAM is efficient even for robot trajectories
with many loops as it avoids unnecessary fill-in in the factor matrix by periodic variable reordering. Also, to enable data   matrix factorization of the smoothing information matrix in association in real-time; Kaess \etal  \mbox{ }suggest efficient algorithms to access the estimation uncertainties of interest based on the factored information matrix. 

We have adopted a total SLAM schema such as iSAM because the information of each line can be updated when necessary. It can be accomplished because the robot position where a given measurement was taken is known. Following section describes how to use geometric information to improve data association.

\section{Association Problem}\label{sec:Association}
The association problem consists on determine where two measurements acquired at different times were originated from the same physical object. In the context of feature--based SLAM, the system must determine the correct correspondences between a recently measured feature and map landmarks. This problem is a challenging task because features are often indistinguishable. The solution of this problem is essential for consistent map construction since any single false matching may invalidate the entire process\cite{Durrant-WhyteMTBS01}.

The direct solution consists of associating a measurement with the closest predicted observation --A predicted observation is a function of the best robot pose and a map feature. This solution, known as \textit{Single Compatibility Test} (SCT), commonly uses the Mahalanobis distance or \textit{Normalized Innovation Squared} (NIS)\cite{Zhang92,baileythesis}. Single compatibility ignores that measurement prediction errors are correlated; hence, it is susceptible to accept incorrect matchings. 

Neira and Tard\'os \cite{neira2001} propose the \textit{Joint Compatibility Test} (JCT); an implementation of the JCT known as the \textit{Joint Compatibility Branch and Bound} (JCBB) algorithm, generates tentative sets of associations and searches for the largest set that satisfies joint compatibility. For a given set of association pairs, joint compatibility is determined by calculating a single joint NIS gate. The benefit of joint compatibility is that it preserves the correlation information within the set of observations and predicted observations.

Note that a validation gates based on the NIS distance (such as SCT and JCT) provides no statistical measure for the rejection of false associations \cite{baileythesis}. 
To overcome false association, the following section introduces a technique that uses both the compatibility test and geometric constraints to find correct matchings between lines.


\subsection{Validation gate based on segments}
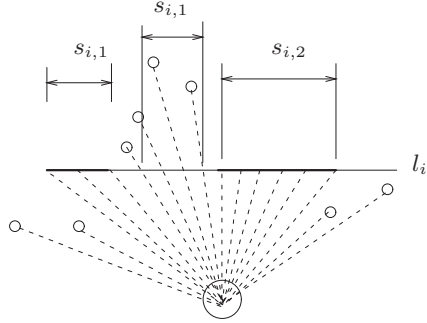
\begin{figure}[!tb]
\centering
\setlength{\unitlength}{0.00045in}
\begingroup\makeatletter\ifx\SetFigFontNFSS\undefined%
\gdef\SetFigFontNFSS#1#2#3#4#5{%
  \reset@font\fontsize{#1}{#2pt}%
  \fontfamily{#3}\fontseries{#4}\fontshape{#5}%
  \selectfont}%
\fi\endgroup%
{\renewcommand{\dashlinestretch}{30}
\begin{picture}(4908,3729)(0,-10)
\put(2524,233){\ellipse{450}{450}}
\put(2160,2754){\ellipse{128}{128}}
\put(1710,3043){\ellipse{128}{128}}
\put(1530,2394){\ellipse{128}{128}}
\put(1395,2034){\ellipse{128}{128}}
\put(3806,1269){\ellipse{128}{128}}
\put(4481,1539){\ellipse{128}{128}}
\put(830,1103){\ellipse{128}{128}}
\put(72,1102){\ellipse{128}{128}}
\thicklines
\path(450,1764)(1170,1764)
\thinlines
\path(450,2979)(450,2574)
\path(1215,3024)(1215,2574)
\path(450,1764)(4590,1764)
\dashline{60.000}(3870,1764)(2570,198)
\thicklines
\path(2475,1764)(3870,1764)
\thinlines
\dashline{60.000}(3510,1764)(2544,219)
\dashline{60.000}(3240,1764)(2535,202)
\dashline{60.000}(2970,1764)(2531,185)
\dashline{60.000}(2745,1764)(2521,189)
\dashline{60.000}(1395,2034)(2496,264)
\dashline{60.000}(1575,2304)(2505,247)
\dashline{60.000}(1710,2979)(2509,230)
\dashline{60.000}(2160,2709)(2519,234)
\dashline{60.000}(2520,1719)(2520,234)
\dashline{60.000}(3780,1269)(2600,215)
\dashline{60.000}(4455,1494)(2593,223)
\dashline{60.000}(1194,1770)(2531,170)
\dashline{60.000}(810,1764)(2475,203)
\dashline{60.000}(495,1719)(2529,168)
\dashline{60.000}(855,1044)(2499,168)
\dashline{60.000}(124,1079)(2507,149)
\path(3870,3024)(3870,1899)
\path(2295,3519)(2295,1854)
\path(1575,3474)(1575,1854)
\path(2520,3024)(2520,1899)
\path(570.000,2829.000)(450.000,2799.000)(570.000,2769.000)
\path(450,2799)(1215,2799)
\path(1095.000,2769.000)(1215.000,2799.000)(1095.000,2829.000)
\path(1695.000,3369.000)(1575.000,3339.000)(1695.000,3309.000)
\path(1575,3339)(2295,3339)
\path(2175.000,3309.000)(2295.000,3339.000)(2175.000,3369.000)
\path(2640.000,2874.000)(2520.000,2844.000)(2640.000,2814.000)
\path(2520,2844)(3870,2844)
\path(3750.000,2814.000)(3870.000,2844.000)(3750.000,2874.000)
\put(4770,1719){\makebox(0,0)[lb]{$l_i$}}
\put(1710,3564){\makebox(0,0)[lb]{$\bar{s}_{i,1}$}}
\put(3105,3069){\makebox(0,0)[lb]{$s_{i,2}$}}
\put(765,3069){\makebox(0,0)[lb]{$s_{i,1}$}}
\end{picture}
}
\caption{An illustration of the segments $S_i = \set{s_{i,1}, s_{i_2}}$ and free-segments $\bar{S}_i = \set{\bar{s}_{i,1}}$ of a line $l_i$.}
\label{fig:reach}
\end{figure}

Let us define $S_i$ the set of segments of the $i$--th line $l_i$ and $\bar{S}_i$ the set of free--segments (the intervals of a line where there is a high probability of free space). The segments $S_i$ are calculated with the inliers of a line, while the free-segments $\hat{S}_i$ are calculated with points such as their measurement ray cross the line $l_i$, see Figure \ref{fig:reach}. The set of segments from a set of points can be obtained straightforward, altough general techniques can be used \cite{Caste-ASME96,Castro_featureextraction}.



To perform geometric association based on segments, the $i$--th line $l_i$ and the new line $l'_j$ are transformed into the same coordinate frame; this transformation allow to treat the segments of a line as a set of intervals. Then the intersection of two segments $A \wedge B$ can be easily calculated by finding those segments that are present both in A and B. The probability that two lines represent geometrically the same line is%
\begin{equation}
\Pr(G | S_i, S_i, S'_j) = \begin{cases}
                \frac{ \norma{S'_j \wedge S_i}}{ \norma{S'_j \wedge S_i} + \norma{S'_j \wedge \bar{S}_i }} & \text{if  }  (\norma{S'_j \wedge S_i} + \norma{S'_j \wedge \bar{S}_i}) \neq 0, \\
		\frac{1}{2}& \text{otherwise,}
               \end{cases}
\end{equation}%
where $\norma{\cdot}$ is the total length of a segment set. When $(\norma{S'_j \wedge S_i} + \norma{S'_j \wedge \bar{S}_i}) = 0$ the local segments are never seen before, and we assign a $0.5$ probability. 

The Segment Validation (SV) gate can be used to improve the Joint Compatibility test. That is, a pair of lines are used in the Joint Compatibility test only if  $\Pr(G | S_i, S_i, S'_j)  \geq 0.5$.


\section{Experimental Results}\label{sec:results}%

\begin{figure}[tbh!]
\centering
\setlength{\unitlength}{0.00017in}
\begingroup\makeatletter\ifx\SetFigFontNFSS\undefined%
\gdef\SetFigFontNFSS#1#2#3#4#5{%
  \reset@font\fontsize{#1}{#2pt}%
  \fontfamily{#3}\fontseries{#4}\fontshape{#5}%
  \selectfont}%
\fi\endgroup%
{\renewcommand{\dashlinestretch}{30}
\begin{picture}(8484,7419)(0,-10)
\path(12,5727)(12,5817)(2082,5817)
	(2082,5727)(12,5727)
\path(8472,102)(8382,102)(8382,7392)
	(8472,7392)(8472,102)
\path(102,102)(12,102)(12,7392)
	(102,7392)(102,102)
\path(4917,102)(4827,102)(4827,1947)
	(4917,1947)(4917,102)
\path(2667,2712)(2667,2802)(4962,2802)
	(4962,2712)(2667,2712)
\path(2757,102)(2667,102)(2667,2802)
	(2757,2802)(2757,102)
\path(12,1362)(12,1452)(1632,1452)
	(1632,1362)(12,1362)
\path(102,102)(102,192)(8382,192)
	(8382,102)(102,102)
\path(8382,327)(8382,12)(8157,12)
	(8157,327)(8382,327)
\path(4962,2712)(4872,2712)(4872,5862)
	(4962,5862)(4962,2712)
\path(3342,5772)(3342,5862)(4962,5862)
	(4962,5772)(3342,5772)
\path(1587,3612)(1497,3612)(1497,5817)
	(1587,5817)(1587,3612)
\path(8472,7392)(8472,7302)(12,7302)
	(12,7392)(8472,7392)
\path(3432,2712)(3342,2712)(3342,4962)
	(3432,4962)(3432,2712)
\path(3342,5727)(4107,5727)
\path(4197,2667)(4962,2667)
\path(2082,5727)(1992,5727)(1992,6537)
	(2082,6537)(2082,5727)
\path(912,2892)(1677,2892)
\path(2622,2802)(2622,1947)
\path(1362,7257)(2127,7257)
\path(8382,6672)(8382,6357)(8157,6357)
	(8157,6672)(8382,6672)
\path(237,6807)(237,6492)(12,6492)
	(12,6807)(237,6807)
\path(8382,3522)(8382,3207)(8157,3207)
	(8157,3522)(8382,3522)
\path(237,3432)(237,3117)(12,3117)
	(12,3432)(237,3432)
\path(12,2757)(12,2847)(1677,2847)
	(1677,2757)(12,2757)
\end{picture}
}
\caption{Synthetic environment used for the tests} \label{fig:objectiveMap}
\end{figure}
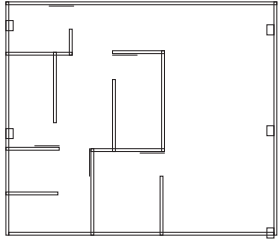

\begin{figure}[tb]
\centering
\includegraphics[scale=0.35]{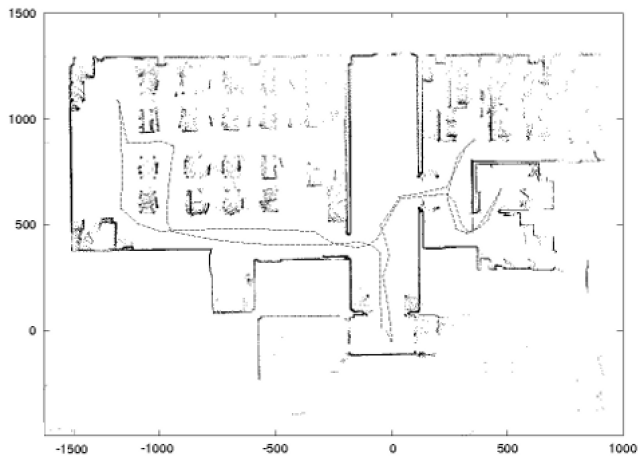}
\caption{Raw data registered with ICP + Lorenztian \cite{romero05} and route (dotted line), the scale is in cm.}
\label{fig:objectiveMapReal}
\end{figure}

To test the ideas presented in this paper, two environments were used: a simulated environment and a real environment shown in figures \ref{fig:objectiveMap} and \ref{fig:objectiveMapReal}, respectively.

\subsection{Simulated Environment}
The simulated environment shown in Figure \ref{fig:objectiveMap} was used to test the ORT performance to find lines from lasers scans; this environment has $42$ lines, some of them are parallel and very close one another ($300 \mathrm{mm}$ between parallel lines representing a door and its corresponding wall); the robot takes $1000$ laser scans from random poses. Laser measurements are corrupted with gaussian noise with $\sigma = 10 \mathrm{mm}$. To find lines from laser scans three algorithms were selected: the sequential Ransac (SR), Split and Merge (SM) and Line Tracking (LT). 

Table \ref{tab:results} shows the experimental results; the first two columns show the True Positives (TP) and the Not Detected Lines (ND). In general, algorithms that use the Odds Ratio Test (LT+ORT and SR+ORT) increase the True Positives and reduce the proportion of lines not detected. The following two columns show the precision of the algorithms; this indicator shows the advantage of the ORT. Finally, the last column shows the speed\footnote{Tests were performed on a ElliteBook 6930p, 2.40 GHz} of the algorithms --one drawback is the increment of the time complexity when using the ORT.

\begin{table}[t]
\centering
\caption {Comparison of line extraction algorithms from synthetic laser scans}  
\label{tab:results}%
\begin{small}
\begin{center}
\centering
\begin{tabular}{p{1.8cm}|cc|cc|c|}
\cline{2-6}
\mbox{\hspace{1.4cm}}& \hspace*{0.4cm}TP\hspace*{0.4cm}& \hspace{0.4cm}ND\hspace*{0.4cm}&
\hspace*{0.2cm} $\mu_{err_r}$ \hspace*{0.2cm}& \hspace*{0.2cm} $\mu_{err_{\alpha}}$ \hspace*{0.2cm} & \hspace*{0.2cm} speed \hspace*{0.2cm} \\
& \mbox{\%}& \mbox{\%} & $[\mathrm{mm}]$ & $[\mathrm{rad}]$ & $[\mathrm{Hz}]$\\
\hline
\lcl{SR} & $91.29$ &  $21.32$ & $6.56$  & $0.0116$ & $196.54$ \\
\hline
\lcl{SM} & $75.86$ &  $24.13$ & $6.68$  &  $0.0143$ & $327.76$ \\ 
\lcl{SM + ORT} & $95.38$ & $12.70$ & $4.37$ &$0.0062$ & $31.56$ \\ 
\hline
\lcl{LT} & $89.97$ & $18.22$ & $4.92$   & $0.0077$ & $59.63$ \\
\lcl{LT+ORT} & $96.82$ & $13.20$& $3.95$ & $0.0055$  & $17.08$ \\
\hline
\end{tabular}
\end{center}
\end{small}
\end{table} 

\begin{figure}[tbh]
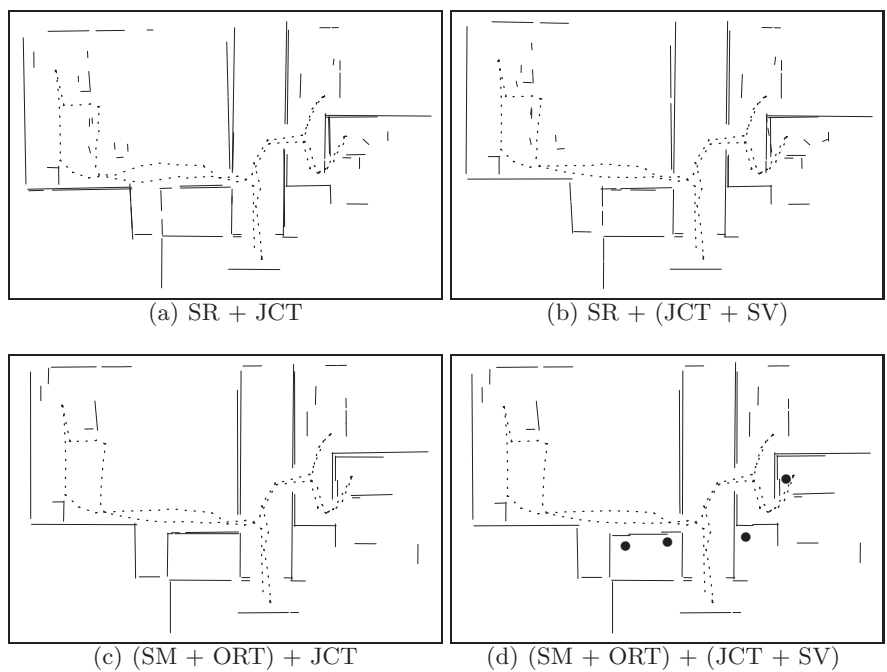

\centering
\subfigure[SR $+$ JCT]{\fbox{\input{testSR.tex}}}\mbox{ }%
\subfigure[SR $+$ (JCT $+$ SV)]{\fbox{\input{testSRSegm.tex}}}
\subfigure[(SM $+$ ORT) $+$ JCT]{\fbox{\setlength{\unitlength}{0.000099in}
\begin{picture}(22864.6,15542.1)(-13584.8,-2473.41)
\drawline(1637.43,663.941)(1684.71,5712.58)
\drawline(1697.68,7097.51)(1753.3,13036.2)
\drawline(-1265.93,12650.3)(-1309.09,4440.61)
\drawline(-1314.87,3342.6)(-1327.96,853.301)
\drawline(2429.05,617.185)(1591.06,622.093)
\drawline(-780.765,635.987)(-1238.56,638.668)
\drawline(-1838.39,642.182)(-5356.24,662.788)
\drawline(4784.63,12967.1)(4283.11,12964.2)
\drawline(4231,12963.9)(3582.54,12960)
\drawline(-85.2791,12938.3)(-1160.82,12931.9)
\drawline(-7543.18,12894.1)(-8273.51,12889.8)
\drawline(-8287.16,12889.7)(-9237.06,12884.1)
\drawline(-9555.39,12882.2)(-12128.8,12866.9)
\drawline(-11409.1,5198.49)(-11434.7,4031.18)
\drawline(-5496.05,3323.53)(-5171.44,3323.56)
\drawline(-5048.81,3323.57)(-1400.65,3323.89)
\drawline(-5245.92,3377.14)(-4775.43,3383.04)
\drawline(-4438.51,3387.27)(-1405.78,3425.35)
\drawline(-7355.15,3921.15)(-7301.2,780.973)
\drawline(-13317.9,12588)(-13312.2,4190.17)
\drawline(-13262.1,3863.92)(-7231.59,3836.48)
\drawline(1772.1,3795.51)(4185.85,3784.53)
\drawline(-7100.39,857.337)(-5930.21,848.281)
\drawline(-1227.91,811.888)(-769.497,808.341)
\drawline(1207.53,793.04)(1695.77,789.261)
\drawline(-9731.83,9327.08)(-10212.9,9314.82)
\drawline(-1451.32,4416.14)(-1477.22,11549.4)
\drawline(-9641.47,10925.5)(-9487.1,9256.05)
\drawline(-12287.7,12805.1)(-12301.7,11908.6)
\drawline(-12734.4,11776.5)(-12734.6,10933)
\drawline(-5487.51,888.174)(-5463.07,3360.41)
\drawline(-12052.3,5137.77)(-11407.4,5151.95)
\drawline(5041.78,5513.6)(7392.99,5565.3)
\drawline(3974.16,5305.58)(3975.96,7978.19)
\drawline(-1446.09,-1224.46)(1458.83,-1206.51)
\drawline(1588.24,-1205.71)(2031.37,-1202.97)
\drawline(4770.5,8916.07)(4763.29,9631.4)
\drawline(4763.28,9632.7)(4755.7,10385.1)
\drawline(4754.18,10535.7)(4748.67,11082.8)
\drawline(5259.91,2755.11)(6477.58,2749.93)
\drawline(1804.46,12604.6)(1813.01,7302.33)
\drawline(1815.79,5581.79)(1818.77,3735.61)
\drawline(5544.02,5379.78)(5109.86,5386.08)
\drawline(6893.34,7709.8)(4131.59,7717.26)
\drawline(8862.17,1593.55)(8874.43,2707.4)
\drawline(4147.72,7776.48)(4146.02,6360.65)
\drawline(4143.05,3884.92)(4141.49,2582.9)
\drawline(4264.36,6387.05)(4270.26,5470.63)
\drawline(2530.3,10300.5)(2526.23,8930.5)
\drawline(3949.19,7935.13)(9441.9,8013.63)
\drawline(3028.36,11768.7)(2999.12,10719.5)
\drawline(-5317.23,568.714)(-5317.56,-747.829)
\drawline(-5317.56,-757.352)(-5317.96,-2367.53)
{\dottedline{1000}(0,0)
(-28.0562,523.807)
(-57.735,1047.66)
(-79.0673,1559.33)
(-89.4468,2070.1)
(-92.4696,2586.47)
(-89.0799,3102.38)
(-73.7181,3605.11)
(-96.9086,3601.81)
(-114.501,3600.25)
(-483.995,3941.85)
(-499.436,3926.58)
(-991.361,4098.72)
(-1001.45,4080.02)
(-1487.48,4031.75)
(-1997.42,4004.01)
(-2507.21,3972.39)
(-3001.86,3952.65)
(-3502.3,3946.91)
(-4032.74,3998.36)
(-4568.24,4018.6)
(-5066.19,4004.75)
(-5572.3,3993.9)
(-6110.98,4025.02)
(-6607.52,4077.62)
(-7117.38,4142.77)
(-7618.52,4230.62)
(-8118.37,4328.07)
(-8618.44,4439.82)
(-9118.93,4561.87)
(-9111.24,4566.97)
(-9097.37,4562.13)
(-9308.67,5014.24)
(-9311.82,5001.32)
(-9292.93,5486.02)
(-9269.77,5987.39)
(-9240.66,6486.86)
(-9201.64,6992.89)
(-9153.76,7489.42)
(-9086.05,7984.24)
(-9014.37,8478.71)
(-9034.26,8480.56)
(-9053.44,8481.25)
(-9068.15,8480.49)
(-9548.15,8692.27)
(-9556.49,8662.33)
(-10068.9,8634.71)
(-10576.4,8614.77)
(-11081.6,8603.04)
(-11080.4,8611.73)
(-11068.3,8621.3)
(-11057.2,8627.94)
(-11171.3,9126.85)
(-11275.1,9625.83)
(-11384.9,10129.7)
(-11472.3,10628.8)
(-11491.7,10638)
(-11522,10701.1)
(-11521.2,10622.6)
(-11519,10667.7)
(-11498,10660.8)
(-11490.3,10653.6)
(-11469.4,10647.4)
(-11425.1,10141.9)
(-11384.9,9626.53)
(-11361.2,9115.72)
(-11331.1,8604.96)
(-11315.7,8097.47)
(-11309.1,7584.66)
(-11308.4,7074.48)
(-11318.2,6566.01)
(-11321.3,6051.58)
(-11348.5,5549.08)
(-11334.1,5552.24)
(-11318.5,5557.44)
(-10936.3,5201.68)
(-10567.2,4841.56)
(-10551.3,4857.3)
(-10063.7,4680.31)
(-9581.29,4503.14)
(-9576.39,4524.96)
(-9066.9,4554.04)
(-8554.39,4587.67)
(-8047.32,4611.36)
(-7531.83,4626.8)
(-7016.82,4639.04)
(-6505.82,4657.05)
(-5996.02,4672.02)
(-5492.57,4725.66)
(-4982.39,4720.26)
(-4472.64,4660.83)
(-3957.74,4626.86)
(-3445.23,4577.5)
(-2937.35,4530.6)
(-2938.17,4516.27)
(-2504.5,4229.54)
(-2064.77,3869.9)
(-2072.64,3889.24)
(-1569,3883.39)
(-1063.47,3800.34)
(-1063.19,3830.68)
(-571.322,3955.96)
(-580.651,3983.52)
(-180.613,4303.33)
(-194.55,4322.2)
(32.4485,4779.71)
(256.126,5233.12)
(498.933,5687.72)
(751.398,6122.25)
(766.502,6120.09)
(769.394,6131.45)
(1273.85,6187.87)
(1776.51,6280.53)
(2274.52,6299.99)
(2789.41,6343.55)
(2795.52,6362.62)
(2768.25,6348.16)
(3124.3,5979.94)
(3113.63,5954.66)
(3243.1,5464.66)
(3372.78,4996.61)
(3389.16,4988.37)
(3688.95,4537.92)
(3705.06,4548.56)
(3720.98,4576.78)
(3737.09,4599.06)
(4201.75,4791.08)
(4191.84,4801.99)
(4553.95,5166.26)
(4534.21,5185.04)
(4727.73,5650.49)
(4912.19,6129.33)
(5109.22,6600.44)
(5085.79,6607.28)
(5065.59,6604.95)
(5046.25,6591.82)
(5035.6,6566.96)
(5030.88,6546.98)
(5040.79,6526.75)
(5059.54,6508.1)
(4801.45,6055.41)
(4561.59,5629.76)
(4305.32,5184.14)
(4043.27,4749.37)
(4032.1,4746.59)
(4027.83,4753.31)
(4031.8,4740.59)
(3599.62,4486.98)
(3605.51,4501.1)
(3602.82,4521.78)
(3121.86,4786.33)
(3132.53,4796.79)
(3148.2,4807.28)
(3133.72,4817.11)
(3146.28,4827.37)
(3006.09,5324.03)
(2859.1,5822.07)
(2722.14,6320.82)
(2733.12,6330.87)
(2834.01,6824.05)
(2934.51,7315.33)
(3052.8,7803.98)
(3184.86,8289.4)
(3213.07,8286.96)
(3524.5,8675.15)
(3873.05,9054.61)
(3877.58,9045.02)
(3875.53,9032.12)
(3858.53,9017.99)
(3864.83,8999.22)
(3851.7,9001.45)
(3837.51,8983.55)
(3824.86,8997.49)
(3458.72,8644.75)
(3455.57,8625.16)
(3288.92,8152.85)
(3092.82,7670.56)
(2897.4,7185.51)
(2683.03,6693.89)
(2674.06,6717.93)
(2669.3,6725.75)
(2164.91,6584.96)
(1679.09,6452.29)
(1186.08,6325.38)
(684.438,6213.65)
(690.592,6220.38)
(701.675,6209.97)
(435.068,5771.63)
(168.283,5362.16)
(-104.787,4913.57)
(-88.3336,4913.09)
(-63.2525,4904.95)
(72.1764,4411.87)
(204.023,3909.03)
(314.925,3411.98)
(301.269,3411.56)
(175.097,2919.73)
(37.6107,2428.62)
(60.1288,2420.15)
(139.259,1916.47)
(230.177,1414.41)
(289.851,911.474)
(353.873,401.419)
(408.433,-105.844)
(445.311,-629.771)
(430.504,-626.861)
(418.553,-626.843)
(406.677,-596.464)
(407.238,-580.33)
(419.871,-555.634)
(427.951,-554.711)
(187.077,-116.844)



}\end{picture}}}\mbox{ }%
\subfigure[(SM $+$ ORT) $+$ (JCT $+$ SV) ]{\fbox{\setlength{\unitlength}{0.000099in}
\begin{picture}(22864.6,15542.1)(-13584.8,-2473.41)
\drawline(1651.42,742.749)(1689.22,5754.89)
\drawline(1699.37,7100.1)(1744.76,13120)
\drawline(-1279.34,12640.6)(-1299.81,4445.14)
\drawline(-1306.52,3355.97)(-1321.91,855.766)
\drawline(2432.19,621.448)(1610.52,624.713)
\drawline(-775.328,634.195)(-1237.28,636.031)
\drawline(-1831.93,638.394)(-5356.28,652.401)
\drawline(-85.6831,12940.5)(-1139.07,12933.1)
\drawline(-7578.76,12887.5)(-9250.05,12875.7)
\drawline(-9587.34,12873.3)(-11945.8,12856.6)
\drawline(-11403,5128.18)(-11420.2,4060.79)
\drawline(-4401.2,3288.62)(-2162.68,3294.32)
\drawline(-2508.97,3416.21)(-1393.37,3419.73)
\drawline(-7340.4,3630.27)(-7319.67,2649.39)
\drawline(-7319.43,2638.24)(-7279.52,750.698)
\drawline(-5306.16,3218)(-4299.81,3215.76)
\drawline(-13358.7,12466.3)(-13347.2,10850.2)
\drawline(-13347.1,10843.3)(-13299.3,4147.79)
\drawline(-13247,3776.73)(-7218.76,3804.21)
\drawline(2737.34,3849.58)(4204.86,3856.27)
\drawline(-7069.91,821.245)(-5912.11,817.826)
\drawline(-1215.17,803.953)(-756.75,802.599)
\drawline(1220.33,796.759)(1708.58,795.317)
\drawline(-9751.83,9312.56)(-10232.9,9297.22)
\drawline(-1446.22,4493.75)(-1450.98,5132.42)
\drawline(-1451.08,5145.46)(-1474.68,11537.1)
\drawline(-9671.84,10915.6)(-9507.6,9248.49)
\drawline(-12323.7,12737.8)(-12333.2,11893)
\drawline(-12770.4,11701.3)(-12764.8,10853.5)
\drawline(-5468.02,841.219)(-5452.88,3322.09)
\drawline(-12044.6,5061.38)(-11399.8,5081.12)
\drawline(5045.63,5584.49)(7396.3,5656.45)
\drawline(3983.77,5347.22)(3992.86,7960.95)
\drawline(-1436.63,-1225.03)(1468.28,-1205.71)
\drawline(1601.02,-1204.83)(2040.67,-1201.9)
\drawline(4791.75,8924.54)(4786.87,9678.32)
\drawline(4786.84,9682.79)(4782.56,10344)
\drawline(4781.54,10501.1)(4778.06,11039.7)
\drawline(5276.53,2790.89)(6471.86,2790.35)
\drawline(1790.8,3803.67)(2751.98,3779.93)
\drawline(1838.55,12603.3)(1828.85,7345.8)
\drawline(1825.57,5571.48)(1822.22,3757.88)
\drawline(8880.23,1667.3)(8887.94,2790.26)
\drawline(5549.06,5434.03)(5113.47,5438.06)
\drawline(6881.52,7783.08)(4118.25,7767.43)
\drawline(4137.15,7825.84)(4144.57,6423.78)
\drawline(4157.73,3937.63)(4164.62,2635.62)
\drawline(4261.02,6437.26)(4274.93,5514.44)
\drawline(2562.84,10276.8)(2549.67,8906.85)
\drawline(3966.42,7900.48)(9459.73,7942.02)
\drawline(4816.48,12939)(4314.57,12934.6)
\drawline(4262.86,12934.2)(3614.02,12928.5)
\drawline(3056.93,11738.3)(3025.06,10689.2)
\drawline(-5308.45,566.616)(-5308.5,-749.464)
\drawline(-5308.51,-759.498)(-5308.58,-2369.66)
{\dottedline{1000}(0,0)
(-29.668,524.901)
(-49.2781,1049.88)
(-69.9606,1561.7)
(-80.4932,2072.49)
(-83.7526,2588.85)
(-80.5678,3104.82)
(-65.2902,3607.57)
(-86.8243,3604.25)
(-106.051,3602.59)
(-475.88,3943.5)
(-491.293,3926.37)
(-983.071,4087.43)
(-992.021,4062.51)
(-1475.77,4014.33)
(-1985.26,3977.33)
(-2494.76,3942.91)
(-2990.46,3921.3)
(-3491.73,3912.96)
(-4020.37,3955.69)
(-4544.98,3930.92)
(-5053.43,3947.94)
(-5560.6,3968.3)
(-6099.39,4002.47)
(-6596.1,4052.54)
(-7105.19,4117.14)
(-7606.7,4196.37)
(-8107.04,4288.83)
(-8607.74,4395.99)
(-9108.93,4513.7)
(-9101.27,4519.07)
(-9087.36,4516.38)
(-9301.66,4979.02)
(-9305.11,4978.08)
(-9290.53,5474.84)
(-9269.36,5976.39)
(-9243.57,6476.28)
(-9207.7,6982.64)
(-9162.92,7479.73)
(-9096.94,7975.26)
(-9028.66,8470.34)
(-9050.3,8473.04)
(-9070.02,8474.77)
(-9087.65,8475.03)
(-9566.13,8683.61)
(-9574.06,8655.1)
(-10086.2,8625.95)
(-10594,8604.49)
(-11102,8591.98)
(-11098,8600.94)
(-11085.6,8610.58)
(-11074.5,8617.29)
(-11191.6,9115.94)
(-11298.1,9614.65)
(-11411.4,10117.9)
(-11500.6,10614.7)
(-11519.6,10611)
(-11546,10599.2)
(-11549.6,10587.4)
(-11545.4,10604.1)
(-11526.5,10594.3)
(-11518.9,10584.3)
(-11498,10578.1)
(-11450.5,10072.6)
(-11406.4,9557.61)
(-11376.9,9047.09)
(-11346.1,8536.5)
(-11327.8,8029.12)
(-11317.9,7516.36)
(-11313.8,7006.17)
(-11320,6497.63)
(-11320.2,5983.1)
(-11344.1,5480.13)
(-11329.9,5483.41)
(-11314.4,5489.09)
(-10930.4,5136.78)
(-10560.6,4779.74)
(-10545.1,4796.05)
(-10057.6,4623.22)
(-9574.85,4450.13)
(-9570.06,4472.69)
(-9060.76,4506.75)
(-8548.44,4543.51)
(-8041.93,4570)
(-7526.03,4586.84)
(-7011.02,4600.46)
(-6500.01,4616.14)
(-5990.08,4624.27)
(-5485.31,4633.96)
(-4974.78,4626.43)
(-4466.53,4616.22)
(-3951.79,4595.02)
(-3439.62,4565.46)
(-2931.57,4519.05)
(-2932.31,4505.46)
(-2498,4225.51)
(-2059.38,3944.92)
(-2067.15,3966.75)
(-1560.97,3896.66)
(-1054.04,3801.79)
(-1054.82,3828.72)
(-562.894,3956.82)
(-570.791,3985.61)
(-169.491,4305.45)
(-183.365,4324.73)
(43.0787,4782.37)
(264.454,5235.96)
(507.285,5691.03)
(760.001,6126.35)
(775.408,6127.19)
(782.54,6155.59)
(1288.02,6224.73)
(1788.77,6294.02)
(2285.59,6346.11)
(2800.84,6384.21)
(2807.44,6372.81)
(2780.4,6357.31)
(3135.17,5983.59)
(3125.23,5962.89)
(3258.77,5470.62)
(3381.14,4984.19)
(3396.11,4994.42)
(3693.14,4582.47)
(3709.49,4596.59)
(3725.71,4618.4)
(3741.81,4640.3)
(4198.04,4834.01)
(4196,4854.14)
(4559.02,5220.48)
(4540.32,5238.13)
(4730.27,5711.84)
(4911.71,6187.99)
(5105.64,6659.99)
(5082.21,6666.58)
(5061.89,6664.31)
(5041.84,6652.41)
(5031.06,6635.73)
(5026.41,6617.07)
(5036.07,6602.29)
(5052.14,6583.62)
(4800.7,6129.34)
(4565.65,5686.32)
(4312.48,5237.75)
(4050.58,4802.68)
(4038.39,4814.78)
(4034.77,4823.97)
(4038.64,4811.25)
(3606.17,4554.37)
(3612.01,4566.39)
(3609.4,4576.66)
(3129.94,4780.6)
(3140.74,4789.03)
(3156.47,4797.51)
(3142.05,4805.47)
(3154.66,4813.83)
(3015.68,5309.05)
(2869.91,5805.6)
(2734.18,6302.88)
(2745.31,6311.12)
(2848.77,6802.32)
(2949.35,7291.54)
(3070,7777.5)
(3204.35,8259.7)
(3232.5,8256.64)
(3545.68,8643.37)
(3895.27,9022.14)
(3898.97,9012.32)
(3891.94,8997.23)
(3879.84,8982.19)
(3883.47,8961.53)
(3873.58,8961.91)
(3860.25,8948.41)
(3847.94,8962.61)
(3479.03,8613.91)
(3473.66,8595.64)
(3305.52,8125.86)
(3108.14,7649.47)
(2911.41,7171.72)
(2695.39,6702.05)
(2686.48,6715.05)
(2681.99,6716.77)
(2176.97,6583.34)
(1690.22,6455.14)
(1196.95,6330.81)
(694.409,6245.06)
(698.075,6226.21)
(708.801,6210.9)
(442.376,5768.68)
(176.097,5341.34)
(-96.9852,4900.59)
(-80.405,4898.99)
(-55.3194,4890.95)
(80.1301,4400.99)
(212.211,3907.51)
(323.508,3412.1)
(311.708,3411.95)
(183.959,2919.95)
(46.49,2428.11)
(69.0882,2419.64)
(150.027,1916.18)
(241.586,1414.38)
(299.194,911.848)
(362.799,401.871)
(417.366,-105.361)
(454.337,-629.265)
(439.466,-626.302)
(427.258,-626.289)
(413.752,-595.881)
(414.392,-579.084)
(429.541,-554.298)
(438.601,-547.232)
(196.869,-115.098)
}



\put(2000,2800){$\bullet$}
\put(-2500,2500){$\bullet$}
\put(-4900, 2300){$\bullet$}
\put(4300, 6100){$\bullet$}
\end{picture}}}
\caption{Map of our laboratory using different techniques. Left column shows the results obtained by using the Joint Compatibility Test (JCT) and the right column maps are obtained with the Joint Compatibility Test with Segment Validation (JCT + SV). The dashed polygon represents the robot trajectory, and doors correctly detected are marked by bullets.}
\label{fig:realTestSAM}
\end{figure}

\subsection{Real Environment}
To test the smoothing and mapping framework --including the proposed techniques to line extraction and data association; we use a real robot to take $215$ laser scans from our laboratory using a Sick LMS--200. This sensor is configured to read $30\mathrm{m}$ over a $180^{o}$ arc, with an angle resolution of $0.5^{o}$. Figure \ref{fig:objectiveMapReal} shows the resulting points map from the same information; as can be seen, the environment is highly cluttered. Resulting line maps by using the SAM framework are shown in Figure \ref{fig:realTestSAM}; here, only the results from two algorithms are presented: the Sequential Ransac (SR) and the Split and Merge with Odds Ratio Test (SM + ORT).  Figures \ref{fig:realTestSAM}a and \ref{fig:realTestSAM}c show the results from the Joint Compatibility Test, while figures \ref{fig:realTestSAM}b and \ref{fig:realTestSAM}d show the results from the Joint Compatibility Test with Segment Validation (SV).

As expected, the line maps are less expressive than the map of raw points; Sequential Ransac approach gives poor quality line maps (figures \ref{fig:realTestSAM}a and \ref{fig:realTestSAM}b) --see for example the inaccurate estimation of angles; The Split and Merge  algorithm gives better results (figures \ref{fig:realTestSAM}c and \ref{fig:realTestSAM}d). Line maps are qualitatively better when the Segment Validation scheme is used. Wrong associations may become more evident
in environments with parallel lines that are very close one another: note the correct detection of doors and walls in Figure \ref{fig:realTestSAM}d. The validation based on the segments of each line prevents wrong associations improving the final map.



\section{Conclusions}\label{sec:conclusions}
We have presented an implementation of the Smoothing and Mapping approach for indoor environments. The contributions of this paper are a probabilistic algorithm to find line clusters from laser scan data, and a validation gate based on segments that  improves the data association for line maps.  

The proposed algorithm to extract lines from laser scans uses a probabilistic criterion to merge clusters rather than an \textit{ad hoc} distance metric; the criterion is stated as a ratio of marginal likelihoods. This criterion allows to decide which model must be merged and to output the best number of models. The proposed algorithm only uses the noise model parameters and avoid to use unnecessary thresholds. Experimental results show that the proposed approach works well to find the correct number of lines, increasing the proportion of True Positives and improving their precision. In other hand, the Segment Validation scheme is helpful to improve the data association when parallel lines are present in the environment. As is shown in the real test, the complete SAM framework finds high quality indoor line maps from cluttered environments.


\begin{thebibliography}{10}

\bibitem{thrunProbabilistic}
Thrun, S., Burgard, W., Fox, D.:
\newblock Probabilistic Robotics (Intelligent Robotics and Autonomous Agents).
\newblock {The MIT Press} (2005)

\bibitem{GrunMDL07}
Gr\"{u}nwald, P.D.:
\newblock The Minimum Description Length Principle (Adaptive Computation and
  Machine Learning).
\newblock The MIT Press (2007)

\bibitem{Ying2010}
Yang, M.Y., F\"{o}rstner, W.:
\newblock Plane detection in point cloud data.
\newblock Technical Report TR-IGG-P-2010-01, Department of Photogrammetry
  Institute of Geodesy and Geoinformation University of Bonn (2010)

\bibitem{Akaike:1974}
Akaike, H.:
\newblock A new look at the statistical model identification.
\newblock {IEEE} Transactions on Automatic Control \textbf{19} (1974)  716--723

\bibitem{neira2001}
Neira, J., Tard\'os, J.:
\newblock Data association in stochastic mapping using the joint compatibility
  test.
\newblock IEEE Transactions on Robotics and Automation (2001)

\bibitem{KaessRD08}
Kaess, M., Ranganathan, A., Dellaert, F.:
\newblock isam: Incremental smoothing and mapping.
\newblock IEEE Transactions on Robotics \textbf{24} (2008)  1365--1378

\bibitem{duda1973pca}
Duda, R.O., Hart, P.E., Stork, D.G.:
\newblock {Pattern classification and scene analysis}.
\newblock Wiley New York (1973)

\bibitem{Borges2000}
Borges, G.A.:
\newblock A split-and-merge segmentation algorithm for line extraction in 2-d
  range images.
\newblock In: ICPR '00: Proceedings of the International Conference on Pattern
  Recognition, Washington, DC, USA, IEEE Computer Society (2000)  1441

\bibitem{Siegwart}
Siegwart, R., Nourbakhsh, I.R.:
\newblock Introduction to Autonomous Mobile Robots.
\newblock Bradford Book (2004)

\bibitem{Thrun03IEEETRANS}
Thrun, S., Martin, C., Liu, Y., H{\"a}hnel, D., Emery~Montemerlo, R., Deepayan,
  C., Burgard, W.:
\newblock A real-time expectation maximization algorithm for acquiring
  multi-planar maps of indoor environments with mobile robots.
\newblock IEEE Transactions on Robotics and Automation \textbf{20} (2003)
  433--442

\bibitem{Han2004}
Han, F., Tu, Z., Zhu, S.C.:
\newblock Range image segmentation by an effective jump-diffusion method.
\newblock IEEE Transactions on Pattern Analysis and Machine Intelligence
  \textbf{26} (2004)  1138--1153

\bibitem{BollesF81}
Bolles, R.C., Fischler, M.A.:
\newblock A ransac-based approach to model fitting and its application to
  finding cylinders in range data.
\newblock In: IJCAI. (1981)  637--643

\bibitem{schnabel-2007-efficient}
Schnabel, R., Wahl, R., Klein, R.:
\newblock Efficient ransac for point-cloud shape detection.
\newblock Computer Graphics Forum \textbf{26} (2007)  214--226

\bibitem{HTSurvey}
Leavers, V.F.:
\newblock Which hough transform?
\newblock CVGIP: Image Underst. \textbf{58} (1993)  250--264

\bibitem{nguyen_2007_a_comparison_of}
Nguyen, V., G\"{a}chter, S., Martinelli, A., Tomatis, N., Siegwart, R.:
\newblock {A Comparison of Line Extraction Algorithms using 2D Range Data for
  Indoor Mobile Robotics}.
\newblock Autonomous Robots \textbf{23} (2007)  97--111

\bibitem{Durrant-WhyteMTBS01}
Durrant-Whyte, H.F., Majumder, S., Thrun, S., Battista, M.D., Scheding, S.:
\newblock A bayesian algorithm for simultaneous localisation and map building.
\newblock In: ISRR. (2001)  49--60

\bibitem{Zhang92}
Zhang, Z., Faugeras, O.:
\newblock A 3-d world model builder with a mobile robot.
\newblock \textbf{11} (1992)  269--285

\bibitem{baileythesis}
Bailey, T.:
\newblock Mobile Robot Localisation and Mapping in Extensive Outdoor
  Environments.
\newblock PhD thesis, Australian Centre for Field Robotics, University of
  Sydney (2002)

\bibitem{Caste-ASME96}
Castellanos, J.A., Tard\'os, J.D.:
\newblock Laser-based segmentation and localization for a mobile robot.
\newblock ASME PRESS, New York (1996)  101--108

\bibitem{Castro_featureextraction}
Castro, D., Nunes, U., Ruano, A.:
\newblock Feature extraction for moving objects tracking system in indoor
  environments.
\newblock In: in Proc. 5th IFAC/euron Symposium on Intelligent Autonomous
  Vehicles. (2004)  5--7

\bibitem{romero05}
Romero, L., Arellano, J.J.:
\newblock Robust local localization of a mobile robot using a 2-{D} laser range
  finder.
\newblock In: ENC '05: Proceedings of the Sixth Mexican International
  Conference on Computer Science, Washington, DC, USA, IEEE Computer Society
  (2005)  248--255

\end{thebibliography}
\end{document}